\documentclass{article} % For LaTeX2e
\usepackage{iclr2021_conference,times}

\usepackage{amsmath,amsfonts,bm}

\def\eqref#1{equation~\ref{#1}}
\def\1{\bm{1}}

\DeclareMathAlphabet{\mathsfit}{\encodingdefault}{\sfdefault}{m}{sl}
\SetMathAlphabet{\mathsfit}{bold}{\encodingdefault}{\sfdefault}{bx}{n}

\usepackage{hyperref}
\usepackage{url}
\usepackage{graphicx}
\usepackage{enumerate}

\title{Knowledge accumulating: The general pattern of learning}

\author{Zhuoran Xu \& Hao Liu \\
JD.com, Inc. \\
\texttt{\{xuzhuoran,liuhao163\}@jd.com} \\
}

\iclrfinalcopy % Uncomment for camera-ready version, but NOT for submission.
\begin{document}

\maketitle

\begin{abstract}
Artificial Intelligence has been developed for decades with the achievement of great progress. Recently, deep learning shows its ability to solve many real world problems, e.g. image classification and detection, natural language processing, playing GO. Theoretically speaking, an artificial neural network can fit any function and reinforcement learning can learn from any delayed reward. But in solving real world tasks, we still need to spend a lot of effort to adjust algorithms to fit task unique features. This paper proposes that the reason of this phenomenon is the sparse feedback feature of the nature, and a single algorithm, no matter how we improve it, can only solve dense feedback tasks or specific sparse feedback tasks.
%End-to-end training can solve many kinds of tasks, but there are still many tasks that require knowledge accumulating.
This paper first analyses how sparse feedback affects algorithm perfomance, and then proposes a pattern that explains how to accumulate knowledge to solve sparse feedback problems.
\end{abstract}

\section{Introduction}
Recently, Artificial Intelligence (AI) archives great success in many fileds. For example, current state-of-the-art computer vision algorithms can achieve human accuracy on image classification. The AlphaGO has won the world champion in playing GO. Except those real world tasks, great progress has also been achieved in the fileds of neural architecture search, transfer learning, one-shot learning, etc.

However, there are still many problems that are difficult to sovle. Auto-programming is still a very hard problem. There is a long way to go to sovle real world coding tasks. Applying reinforcement learning into real world still faces many difficulties. In autonomous driving, instead of using reinforcement learning to obtain a end-to-end system, the main methodology separates the whole system into localization, perception, planning and control.

Furthermore, we do not have a uniform theory for Artificial General Intelligence (AGI). Theoretically speaking, by recording inputs and outputs of a human and using supervised learning to fit those data, the intelligence of this human can be fullly imitated, resulting in acquisition of AGI. But we all know this is impossible. Knowledge accumulating is an very important skill of human intelligence, but seldom researches focus on this topic.
%A dream of AI researchers is automatically generating algorithms. Researches like neural architecture search can generate a small set of algorithms. But it is still difficult to use an algorithm to devise a creative and powerful algorithm.
%From the view of learning, learning algorithms are also learnable, that means a learning algorithm is generated by another learning algorithm. 
%Then what is the initial algorithm that generates them all? Innovation and self-examination are common human behavior. Are they unique mechanism generated by evolution or just phenomenon naturally emerged from the complexity of knowledge? So many problems are going to be answered.
Hence, one shortcoming of current AI researches is the lack of a theory about general pattern of learning. This pattern should explain following points:
\begin{enumerate}[(1)]
\item Why research community make progress rapidly in some tasks and slowly in other tasks. What kind of problems can be solved and what can not.
\item For those unsolveable problems, how human solve them?
\item According to the human solution, how to design a artificial solution.
\item How intelligence behaviors like innovation and self-examination emerge in this solution.
\end{enumerate}  

This paper focuses on the sparse feedback phenomenon of the nature. From the sparse feedback phenomenon, this paper believes a single algorithm, no matter how we improve it, can only solve dense feedback tasks or specific sparse feedback tasks. Human have the ability to convert a sparse feedback task into several dense feedback tasks. This ability is generated by thousands of years of evolution and decades of study. This paper also believes that it is inconvincible that all kinds of tasks can be solved in end-to-end training way. Knowledge accumulating plays a key role in human intelligence. So this paper proposes a learning system that can systematically accumulate knowledge, results in the ability of independently solving sparse feedback tasks.

\section{Related work}
\subsection{end-to-end training}
Recently, end-to-end deep learning achieves great progress in many fileds.
In image classification, using deep artificial neural network and end-to-end training, current algorithms \cite{alexnet} \cite{resnet} can achieve human performance in ImageNet dataset.
In object detection, deep learning based algorithms \cite{yolo} \cite{ssd} \cite{fasterrcnn} satisfy the requirements of solving real-world tasks in both speed and accuracy. Focal loss \cite{focalloss} proposes that inbalance of positive and negtive samples has negtive impact on training object detection networks and reweighting loss function can solve this problem. 
In image segmentation, U-Net \cite{unet} proposes to use feature fusion to solve semantic segmentation tasks. Mask R-CNN \cite{maskrcnn} uses Faster R-CNN \cite{fasterrcnn} framework to solve instance segmentation tasks.

Deep metric learning is initially proposed in face recognition and later used by other fileds. Those methods \cite{tripletloss} \cite{centerloss} learn a feature extractor that guarantees objects of the same class having smaller distance in feature space than objects from different classes. The feature extractor has good generalization ability and can be applied to classes that not include in the training set.

Generative adversarial networks (GAN) \cite{gan} can generate new artificial data from the same data distribution of training set. It includes a generative model G that captures the data distribution, and a discriminative model D that estimates the probability that a sample came from the training data rather than G. The GAN training procedure is easy to be unstable. WGAN \cite{wgan} improves the stability of learning and gets rid of problems like mode collapse.

Transformer proposed in \cite{attentionisallyouneed} uses attention mechanism to improve the ability of sequential data processing. Based on the power of Transformer, BERT \cite{BERT} uses self supervised learning and huge text corpus to obtain a pre-trained language representation which largely improves the perfomance of many downstream natural language processing (NLP) tasks.

In reinforcement learning, end-to-end learning algorithms \cite{alphago} can defeat human champions in playing GO. How to play computer games can be easily learned by a single network with images as inputs and actions as outputs \cite{atarigame} \cite{DeepQLearning}. 

\subsection{Meta learning}
Many researches are dedicated to solve meta learning problems. End-to-end training also plays an important role in the recent progress of meta learning.

In transfer learning, it is very popular to use pre-trained models in downstream tasks, e.g. ImageNet pre-trained models in computer vision and BERT in NLP. Paper \cite{transferlearningzhenni} analyses what is being transferred in transfer learning. Cross domain transfer is also available, e.g. paper \cite{clip} learns transferable visual models from natural language.
In increamental learning, paper \cite{rpsnet} proposes to use a supernet that contains a lot of possible execution paths and progressively choose optimal paths for new tasks while encouraging parameter sharing between tasks. In synthetic gradients, paper \cite{Syntheticgradients} proposes a way to generate gradients in back propagation to speed up network training.

In neural architecture search (NAS), paper \cite{neat} proposes to use genetic algorithms to generate network architectures and weights together. In paper \cite{NASrein}, reinforcement learning is used to search network architectures. DARTS \cite{Nasbp} proposes to use back propagation to generate network architectures. Moreover, ENAS \cite{Nasoneshot} introduces the concept of Supernet which is a larger graph containing a lot of possible sub-graphs and the search of weights and sub-graphs are simultaneously done by back propagation.

In one-shot and few-shot Learning, many methods \cite{SiameseNeuralNetwork} \cite{tripletloss} are based on metric learning. Those methods first construct training pairs containing samples from same or different classes, and then optimize a feature extractor to use distance between features to decide if two samples belone to the same class. Matching Network \cite{MatchingNetworks} enrichs the way of constructing training set as N-way K-shot dataset settings. Noticing a fixed distance-metric may not be the optimal choice, Relation Network \cite{RelationNetwork} concatenates features of two samples and uses a network to calculate their similarity.
Task adaptation is another kind of methods. MAML \cite{MAML} proposes a framework to train a meta model on source tasks and fine-tuning it on target tasks. The meta model should be capable of adaptation easily to target tasks. It is done by simulating the adaptation process many times, and optimize the meta model to maximise the adaptation ability across those simulations.

Self supervised learning uses supervised learning and delicate learning strategy to learn feature extractors from unlabled data. Contrastive Predictive Coding \cite{ContrastivePredictiveCoding} proposes that good latent representations should be helpful to predict representations of future samples and discriminative to noise samples. Momentum Contrast \cite{moco} can outperform its supervised pre-training counterpart in many downstream tasks. BYOL \cite{byol} proposes a teacher-student framework that achieves state-of-the-art perfomance without using negtive samples.

\subsection{Limitation of end-to-end training}
Although end-to-end training has achieved so much success, human intervention are still very important in a lot of tasks. In autonomous driving, the using of end-to-end reinforcement training is limited. The most popular framework uses four human designed parts: localization, perception, planning and control \cite{ADSSurvey}. In auto-programming, it is still very difficult for auto generated code pathes to produce correct outputs \cite{pathplausibilityandcorrectness}.

\subsection{Learning system}
Many works are dedicated to design a powerful learning system.
Learning classifier systems \cite{LearningClassifierSystemsreview} \cite{LearningClassifierSystemsrobot} \cite{LearningClassifierSystemscls} are rule-based systems in which rules are generated by genetic algorithms and the execution path of rules are generated by temporal difference learning according to environment.
Gödel machines \cite{Godelmachines0} are self-updating machines that use a recursive self-improvement protocol in which it rewrites its own code when it can prove the new code provides a better strategy.
Ensemble methods can also be considered as a learning system. Paper \cite{CollaborationofExperts} archives 80\% Top-1 Accuracy on ImageNet with 100M FLOPs by generating multiple models in training and selecting the most appropriate one for predicting.
%In contrast to algorithms that focues on a single models, ensemble learning studies how to combine multiple models to better solve problems. 
Neural Turing Machines \cite{NeuralTuringMachines} introduce memory mechanism into neural networks with the ability of fully differentiable and end-to-end trainable. 
Neural Production Systems \cite{neuralproductionsystem} are end-to-end trainable rule systems that use entity properties to match rules and use matched rules to update entity properties.

\section{Problem definition}
\subsection{Sparse feedback}

Sparse feedback is nature phenomenon that widely exists in learning tasks. For example, in autonomous driving, police officer directing traffic is a low possibility event. In this event, the gesture of police officer occupies only a small image area and a few time. Without any priori knowledge, it is very hard for end-to-end algorithms to find police officer's instruction is important for path planning. Most current algorithms depend on smooth and continuous loss functions. But sparse feedback leads to sparse and non-continuous loss functions. Hence, it is difficult to apply current algorithms to sparse feedback tasks. Reinforcement learning has the ability to solve delayed reward tasks. But in practice, constructing a smooth reward function is very important to find a good solution. It is very common in practice to convert a sparse feedback problem into a dense feedback problem. For example, in a simple reinforcement learning task that a model is trained to control a agent to catch a ball. The reward function is usually defined as the distance to the ball instead of a boolen vaule indicates if the ball is reached. So reinforcement learning may also fail in pure sparse feedback environment. This paper believes a single algorithm, no matter how we improve it, can only solve dense feedback tasks or specific sparse feedback tasks. 

Even in dense feadback tasks, many human priori knowledge has been used to imporve perfomance. Paper \cite{FusionPainting} uses fixed segmentation models to largely improve accuracy of point cloud detection. In image captioning \cite{ImageCaptioning0}, it is very common to use a fixed classification or detection models to provide information of image. 
Modifying loss function to adjust or smoothen feedback is a very useful skill in many fileds. Per-class weighted loss is a very common method in image classification which adjusts learning procedure to focus on rare classes. Focal loss \cite{focalloss} balances positive and negtive samples in image detection. WGAN \cite{wgan} builds a smooth loss between discriminator and generator which makes training procedure more stable.
Annotation itself is a behavior that builds dense feedback. The quality of annotation largely affects training results.

\subsection{Task partition and knowledge accumulating}
Human solves sparse feedback problems by using priori knowledge. In the directing traffic task, it is easy to train a model to recognise the gesture of police officers. It is also easy to use the gesture of police officers in path planning. Gesture recognition is a dense feadback task. Using gesture is also a dense feadback task. Both of them can be solved using current algorithms. This paper assumes that all sparse feedback tasks can be separated into several dense feadback sub-tasks.

A solved sub-task is very valuable. Because there are similarities between tasks. The solution of a sub-task can be used directly or with minor modification in others similar tasks. Hence, accumulating solution of sub-tasks, or knowledge accumulating, is very important for solving future tasks.

\subsection{Human tutor}
The key problem is how to separate sparse feedback tasks into dense feadback tasks. This paper assumes that this work can be done by human tutors. A lot of practice works can support this assumption.
For example, image classification tasks are usually separated into parts of preprocessing, augmentation, model structure, loss function. In autonomous driving, a system is separated into parts of localization, perception, planning and control.
Human tutors spend thousands of years of evolution and decades of study to obtain this ability. Hence, this paper does not want to devise an algorithm to implement this ability. We only focus on how to transfer those priori knowledge to machines.

\subsection{Universal algorithms}

Paper \cite{NoFreeLunch} has proven that there is no algorithm that can perform best in all tasks. Our paper wants to select one or several algorithms, called universal algorithms, as foundation of the learning system to solve dense feedback tasks. Those universal algorithms are not the best for a specific dense feedback task, but should have the ability to properly solve most dense feedback tasks.

Current learning algorithms can be categorised into supervised learning, unsupervised learning, half supervised learning, self supervised learning, reinforcement learning. From the view of inductive bias, supervised and reinforcement algorithms learn directly from feedback, so they have weak bias. On the contrary, unsupervised learning, half supervised learning and self supervised learning very depend on human defined task specific bias.
%Furthermore, the design of unsupervised learning, half supervised learning and self supervised learning is a try and error procedure, which depends on the use of feedback. This paper do not use those algorithms as the foundation stone of the learning system. Instead, this paper hopes by using supervised learning and reinforcement learning, the system can generate those algorithms. 
Hence, supervised learning and reinforcement learning are used as universal algorithms for the system.

\subsection{Problem definition}
The problem is defined as below:
\begin{enumerate}[(1)]
\item A system is initialized with several universal algorithms. Universal algorithms can sovle dense feedback tasks, but can not solve sparse feedback tasks.
\item The system faces a serial of tasks. Some of them are dense feedback, some of them are sparse feedback.
\item Those sparse feedback tasks can be separated into dense feedback tasks by human tutors.
\item The system should learn the knowledge of separating tasks from human tutors in the process of solving tasks. Finally, the system should have ability to solve sparse feedback tasks without help of human tutors.
\end{enumerate}

\section{The learning system}
\subsection{pattern of knowledge accumulating}
\begin{figure}[h]
\begin{center}
\includegraphics[scale=0.5]{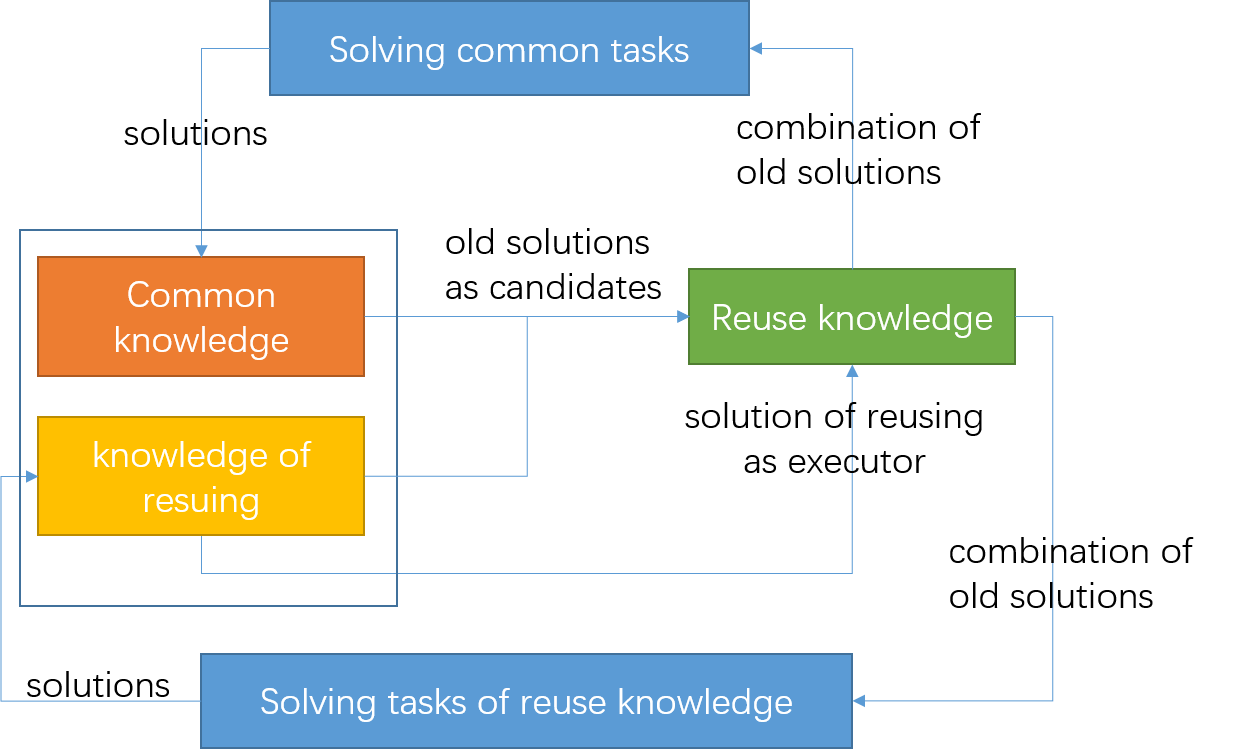}
\end{center}
\caption{Accumulating knowledge and resuing knowledge.}
\label{fig:idea}
\end{figure}

We first concludes the pattern of knowledge accumulating in Figure.\ref{fig:idea}.
At the beginning, solutions of common tasks, refer to image classification or detection .etc, are simply stored. As the purpose of knowledge accumulating is to reuse them to solve new tasks, it is necessary to obtain the knowledge of reusing knowledge, which is done by solving tasks of how to reuse knowledge. Finally, a part of knowledge of resuing works as executor, which takes both common knowledge and knowledge of resuing as inputs and generates solutions for both new common tasks and new tasks of resue knowledge.

Usually, we use an algorithm to learn a task. The solution of the task is learnable, but the algorithm is fixed. In the proposed pattern, both candidates solution and executors, which are used to solve new tasks, are not fixed. They are learned from other tasks. Moreover, it is possible that the executor and the solution generated by it share some same knowledge. This pattern maximises the flexibility of the system, but requires a new architecture to support it.

\subsection{system details}
\begin{figure}[h]
\begin{center}
\includegraphics[scale=0.5]{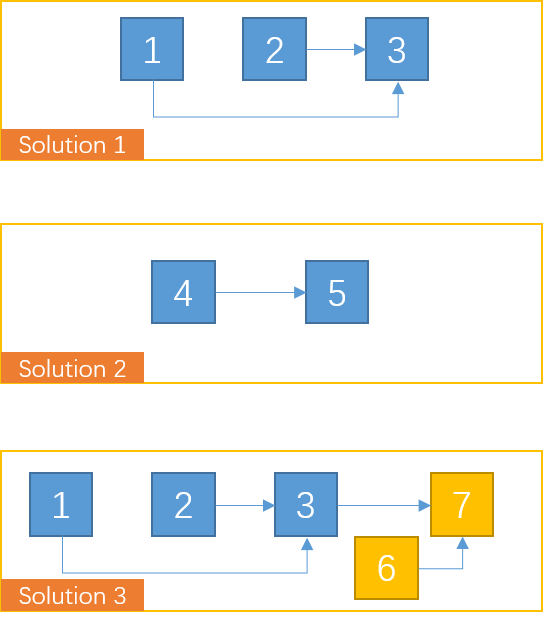}
\end{center}
\caption{Models, dependency and solutions.}
\label{fig:models}
\end{figure}

The whole system works as below:

1. The system are required to solve a serial of tasks. When solving a specific task, human tutors decide if the task needs partition. If partition is not needed, a model is trained using the universal algorithms. If the partition is needed, models are trained for each sub-tasks, in which some sub-tasks are responsible for combining outputs of other sub-tasks. Hence, each task generates a serial of models and dependency between them. Formally, the solution of a task is defined as a tuple $\langle M, D\rangle$, where $M$ is a set of models, $D$ is the dependency between those models. This architecture is shown in Figure.\ref{fig:models}. The models 4 and 5 are unique models for solution 2. Solution 3 shares some models with solution 1. The models 6 and 7 are unique models for solution 3.

\begin{figure}[h]
\begin{center}
\includegraphics[scale=0.5]{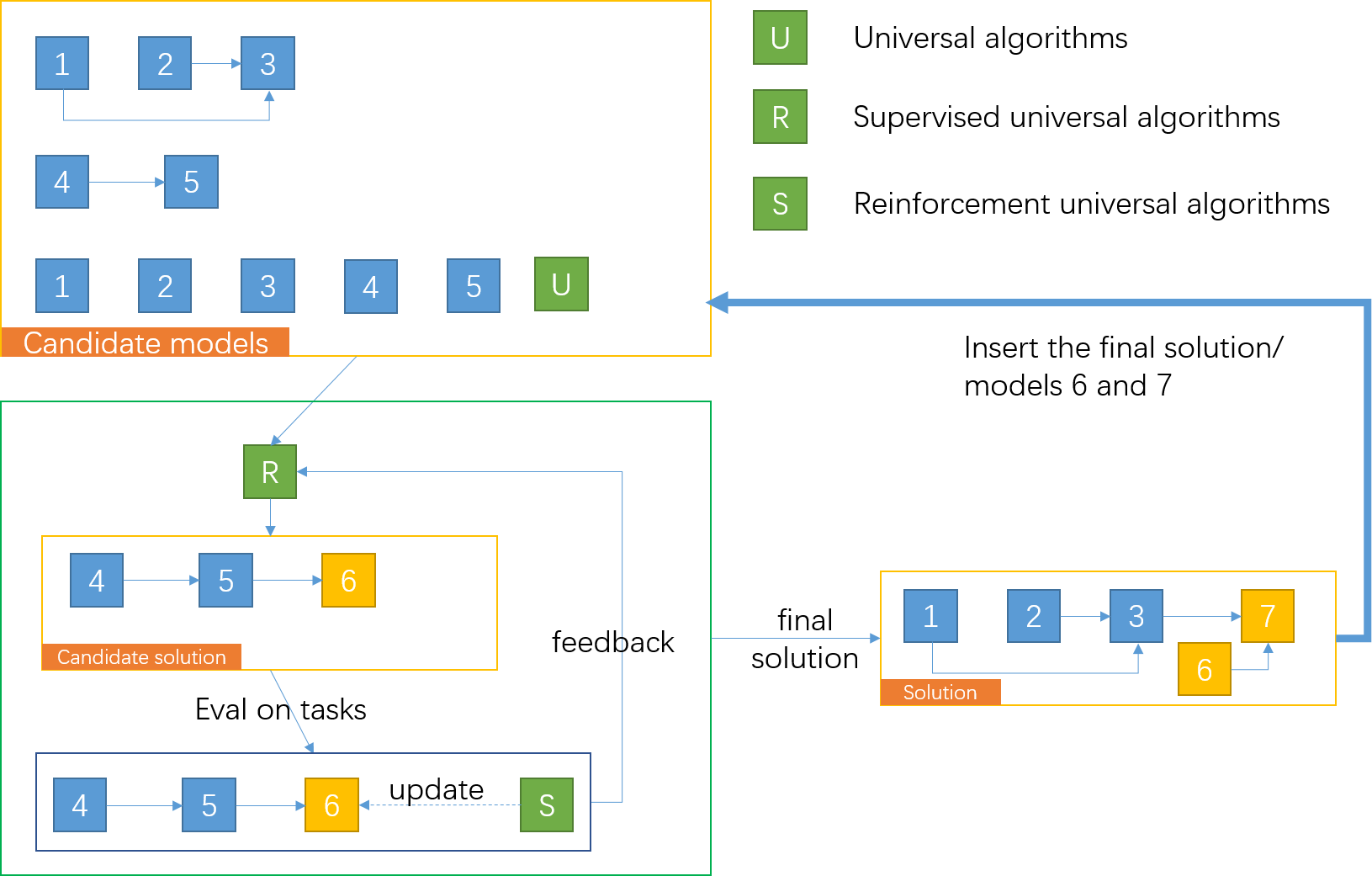}
\end{center}
\caption{Reusing old models.}
\label{fig:reusemodels}
\end{figure}

2. After solving a number of tasks, new tasks usually have similarity with old tasks. So it is helpful to reuse old models to solve new tasks. There are two steps to resue old models to solve a new tasks. First, using universal algorithms, like reinforcement learning, to choose useful old models and generate their dependency. The weights of the chosen models are fixed in new tasks. Second, using universal algorithms, like supervised learning, to train the newly added models to finish the task. A reusing process is shown in Figure.\ref{fig:reusemodels}, where a reinforcement learning algorithm takes old models and all universal algorithms as inputs and generates candidate solutions. The candidate solutions are evaluated in the given task and rewards are passed to the reinforcement learning algorithm for generating better candidates in the next iteration. A supervised learning algorithm is used to optimize new models in the candidate solutions in order to get evaluation results. The final solution and newly added models are inserted into the set of candidates models for resuing in future tasks.
    
\begin{figure}[h]
\begin{center}
\includegraphics[scale=0.5]{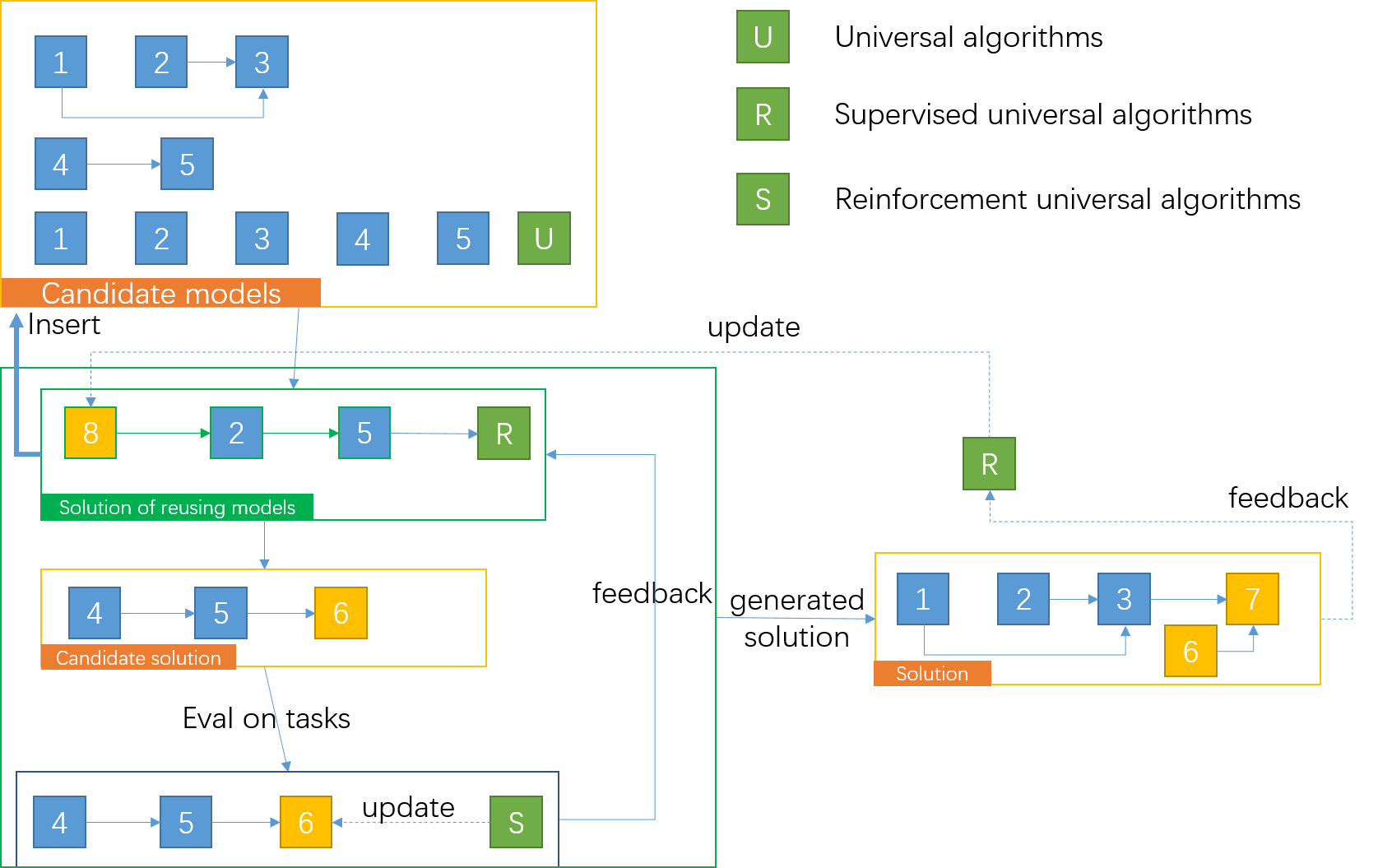}
\end{center}
\caption{Train a solution to reuse old models.}
\label{fig:learntoreuse}
\end{figure}

3. Furthermore, the reinforcement learning algorithms in the first step can be replaced by learned solutions. The very important thing is how to resue old models to solve new tasks is also a task. If it is a sparse feedback task, human tutors need to do partition. So the solution of this kind of tasks is also a tuple $\langle M, D\rangle$. Models obtained in reusing model tasks can also be reused to solve new tasks. Any old model, no matter in what task it is trained, can be used as candidates for solving new tasks. Universal algorithms or any human designed algorithms can also be used as candidates for solving new tasks. Moreover, the above tasks can be upgraded as how to reuse old models to solve new tasks according to feedback. The solution of this kind of tasks is very similar to what we called learning algorithms. The whole process is explained in Figure.\ref{fig:learntoreuse}, where the solution of reusing models is responsible for generating suitable models dependency for a given task. This solution has three models and a reinforcement learning algorithm as a part of it. The models 2 and 5 are old models and fixed in this solution. The model 8 is newly introduced in this solution, which is updated by a reinforcement learning algorithm according to the perfomance of the generated solution. We can use the reinforcement learning for updating here because the solution is separated into four parts and the update of model 8 is assumed as a dense feedback problem. 
It is not enough to update model 8 only on a single task. The mode 8 should be optimized on a serial of tasks. Finally, the model 8 and the solution of resuing is inserted into candidates of models.

4. How to separate sparse feedback tasks is also a task (called partition task), which can be solved in the above way. Furthermore, how to reuse old models to sovle new partition tasks is also a task and can also be sovled in the above way. 
Because many sparse feedback tasks may share the same partition way and models generated in solving a sparse feedback task represent knowledge of partitions, it is possible to combine old models to solve new sparse feedback tasks instead of doing partition by human tutors. So after solving a number of those tasks, the system can generate a closure of learning. 

\subsection{System Explanation}
As deep learning is driven by a huge mount of data, the proposed system is driven by a huge amount of tasks. Those tasks include real-world tasks like image classification, detection, and abstract tasks like reusing models. Everytime a sparse feedback task is solved, some priori knowledge of human tutors is learned by the system. The priori knowledge is fixed in models and dependency between those models. After solving a number of tasks, adequate priori knowledge is learned by the system, especially knowledge of how to reuse old models. With those old models and solutions of how to resue models, any task can be converted into dense feedback sub-tasks by the system.

The solution of each task has a set of models and dependency between them. Those models and dependency may share across solutions. When reusing a old model in a solution, this old model may depend on other old models. Hence, there is a complex dependency hierarchy between models and solutions. Although there are different kinds of tasks, there is no boundary for model reusing. A task can use models generated by any kind of tasks. Moreover, there is no need to keep the system differentiable, so any human designed algorithm can be used in a solution.
%The system is also a self-updating system. Everytime a task is solved, new models and dependency are stored, and can be reused in next task.
%When reusing a model that depends on other models, the system presents a hierarchical structure. Reusing old models makes the system have the ability to solve new tasks without human tutors.

We compare the proposed system with current researches and list several important points below.
\begin{enumerate}[(1)]
\item Current researches focus on solving each task independently. For example, we usually train different models for different tasks in image detection. Although using pre-trained ImageNet models is very popular, it is rare to directly use outputs of a model as inputs of another model. On the contrary, this paper believes that the number of tasks and relationship between tasks are very important. Accumulating models of different tasks and reusing them to solve new tasks are key to artificial general intelligence.

\item Current researches focus on solving real world tasks. Even NAS and meta-learning solve problems originated from real world tasks. This paper proposes that there are many tasks which do not exist unitl we accumulate a lot of models. Those emerged tasks, which can be called as abstract tasks, are very important, especially the tasks of how to reuse old models.

\item Current researches focus on proposing new algorithms. But this paper believes systematically accumulating models of different tasks is equivalently important. The amount of tasks is very important to artificial general intelligence.

\item End-to-end training is widely used in current researches. This paper believes that there is no conflict between end-to-end training and partition of tasks. Because there is no need to separate a dense feedback task that can be learned easily in end-to-end way.

\item Researches like \cite{Capsules} use paths of neurons to construct different functions. This paper believes the interpetability of models is much more clear than neurons. So models are used as building blocks of the system. The paths in paper \cite{Capsules} are determined by a routing-by-agreement mechanism. But in this paper, the ability of generating paths is learned from tasks. Researches of intelligent agent \cite{Tu1999ArtificialAF} design mechanism to determine what is the task like now and switch between tasks. As the purpose of this paper is only to describe the pattern of learning, this kind of mechanism is not provided. The selection of a task to run is determined by human.
\end{enumerate}
%\item For multi-tasks, current researches try to solve them using only one model. Ensemble learning is an exception, which uses multiple sub-models to solve tasks. This paper goes further. It is allowed for the proposed system to use entirely different models to sovle different tasks. 

\section{discussion}
\subsection{knowledge accumulating}
Human spend decades on accumulating knowledge and studying how to use knowledge to solve new tasks. Only accumulating knowledge, but not knowing how to reuse them does not have any positive effect on intelligence. Beware of this, the main mechanism of the proposed system are knowledge accumulating and knowledge reusing.

\subsection{Gödel’s Incompleteness Theorems}
A concern is wehther the universal algorithms and ability of human tutors affect the upper-bound of this system. Does the system have the ability to solve problems that the universal algorithms and human tutors can not solve? We think the answer is yes.
Because tasks are the most important elements in learning. If the system solves tasks that human tutors has never met, the system will outperform human tutors.
The universal algorithms do affect the perfomance of dense feedback tasks. But we believe it is not a critical element for intelligence. Because it is common that different people have different ability on a specific task.

\subsection{Self-update}
Another concern is wehther the system can self-update itself. As mentioned before, the system is designed to have this ability. The system is orgnized by a serial of independent solutions. Each solution only depends on a part of of models of the whole system. An existing model is not allowed to update, because the update will confuse models and solutions depending on it. But the solution of a task can be replaced by a newer one. It is also possible for one part of the system to update another part of the system. For example, solution A, depends on a set of models, can update dependency of another solution. And solution B, depends on another set of models, can update solution A. 

\subsection{Innovation and self-examination}
Due to the complex dependency between models, there is naturely many requirements of continuously optimizing the solutions. For example, after solving a task, the newly generated models may be helpful to sovle a old task in a better way. When the system is very complex, it is also hard to guarantee the most suitable old models are used to sovle a new task. Hence, there are nature requirements to do innovation or self-examination across the system. The system can implement innovation or self-examination by occasionally trying different solutions for a solved task.

\subsection{Root algorithms}
As described in paper \cite{NoFreeLunch}, there is no super algorithm that can perform best in all tasks. But is there an algorithm that can generate all other algorithms? From the system described in this paper, the key is to build a knowledge accumulating system. The universal algorithms of this system can be seen as root algorithms.

\subsection{Create other algorithms}
A dream of AI researchers is automatically generating algorithms. Researches like NAS can generate a small set of algorithms. But it is still difficult to use an algorithm to devise a fully creative algorithm.
From the system described in this paper, the way of reusing old models is very similar to what we called learning algorithms. Because a solution of reusing old models takes feedback as input and adjusts the combination of old models multiple times to maximise feedback.
%Those solutions can be used to create unsupervised learning, half supervised learning, self supervised learning algorithms. Because those kinds of algorithms use features of a specific task to accelerate task solving, which is very similar to find good old models to accelerate task solving.
However, those solutions are mainly focused on solving sparse feedback tasks. It is low proprity to create a solution to solve dense feadback tasks.

\section{Roadmap to Artificial General Intelligence}
Based on the assumption of sparse feedback, a roadmap to artificial general intelligence is presented here:
\begin{enumerate}[(1)]
    \item In this stage, researches focus on solving independent dense feedback tasks (and some specific sparse feedback tasks). Besides algorithms, this stage also accumulates dense feedback data. Finally, a number of algorithms for solving dense feedback tasks and many model instances are obtained.

    \item In this stage, models and algorithms obtained from the first stage can be used as old models and universal algorithms. When facing new tasks, the old models are manually reused or reused by running universal algorithms. A database is used to track how human tutors and universal algorithms reuse those old models.

    \item In this stage, the databse obtained in the second stage is used to train models that learn how to reuse old models. Simulation systems may be required to systematically learn reusing ability.

    \item In this stage, many requirements naturally emerge from the complexity of the system, like innovation and self-examination. After solving those tasks, the system becomes more intelligence.
\end{enumerate}

\section{Conclusion}
This paper proposes that a single algorithm, no matter how we improve it, can only solve dense feedback tasks or specific sparse feedback tasks. Knowledge accumulating is the key for solving sparse feedback tasks. Human tutors can sovle sparse feedback tasks based on their priori knowledge. This paper proposes a learning system which accumulates those priori knowledge and finally obtains the ability to solve sparse feedback tasks without human tutors.

\bibliography{iclr2021_conference}
\bibliographystyle{iclr2021_conference}

\end{document}